# Augmented Curation of Unstructured Clinical Notes from a Massive EHR System Reveals Specific Phenotypic Signature of Impending COVID-19 Diagnosis

FNU Shweta[1]*, Karthik Murugadoss[2]*, Samir Awasthi[2], AJ Venkatakrishnan[2], Arjun Puranik[2], Martin Kang[2],
Brian W. Pickering[1], John C. O'Horo[1], Philippe R. Bauer[1], Raymund R. Razonable[1], Paschalis Vergidis[1],
Zelalem Temesgen[1], Stacey Rizza[1], Maryam Mahmood[1], Walter R. Wilson[1], Douglas Challener[1],
Praveen Anand[2], Matt Liebers[2], Zainab Doctor[2], Eli Silvert[2], Hugo Solomon[2], Tyler Wagner[2],
Gregory J. Gores[1], Amy W. Williams[1], John Halamka[1], Venky Soundararajan[2+], Andrew D. Badley[1+]

[1] Mayo Clinic, Rochester MN, USA
[2] nference, Cambridge MA, USA
* Joint first authors
+ Address correspondence to: ADB (Badley.Andrew@mayo.edu), VS (venky@nference.net)

**Understanding the temporal dynamics of COVID-19 patient phenotypes is necessary to derive fine-grained resolution of pathophysiology. Here we use state-of-the-art deep neural networks over an institution-wide machine intelligence platform for the augmented curation of 15.8 million clinical notes from 30,494 patients subjected to COVID-19 PCR diagnostic testing. By contrasting the Electronic Health Record (EHR)-derived clinical phenotypes of COVID-19-positive (COVID$_{pos}$, n=635) versus COVID-19-negative (COVID$_{neg}$, n=29,859) patients over each day of the week preceding the PCR testing date, we identify anosmia/dysgeusia (37.4-fold), myalgia/arthralgia (2.6-fold), diarrhea (2.2-fold), fever/chills (2.1-fold), respiratory difficulty (1.9-fold), and cough (1.8-fold) as significantly amplified in COVID$_{pos}$ over COVID$_{neg}$ patients. The specific combination of cough and diarrhea has a 3.2-fold amplification in COVID$_{pos}$ patients during the week prior to PCR testing, and along with anosmia/dysgeusia, constitutes the earliest EHR-derived signature of COVID-19 (4-7 days prior to typical PCR testing date). This study introduces an *Augmented Intelligence* platform for the real-time synthesis of institutional knowledge captured in EHRs. The platform holds tremendous potential for scaling up curation throughput, with minimal need for retraining underlying neural networks, thus promising EHR-powered early diagnosis for a broad spectrum of diseases.**

Coronavirus disease 2019 (COVID-19) is a respiratory infection caused by the novel Severe Acute Respiratory Syndrome coronavirus-2 (SARS-CoV-2). As of April 28, 2020, according to WHO there have been nearly 3 million confirmed cases worldwide and more than 200,00 deaths attributable to COVID-19 (https://covid19.who.int/). The clinical course and prognosis of patients with COVID-19 varies substantially, even among patients with similar age and comorbidities[1]. Following exposure and initial infection with SARS-CoV-2, likely through the upper respiratory tract, patients can remain asymptomatic although active viral replication may be present for weeks before symptoms manifest[1,2]. The asymptomatic nature of initial SARS-CoV-2 infection in the majority of patients may be exacerbating the rampant community transmission observed[3]. It remains unknown which patients become symptomatic, and in those who do, the timeline of symptoms remains poorly characterized and non-specific. Symptoms may include fever, fatigue, myalgias, loss of appetite, loss of smell (anosmia), and altered sense of taste, in addition to the respiratory symptoms of dry cough, dyspnea, sore throat, and rhinorrhea, and well as gastrointestinal symptoms of diarrhea, nausea, and abdominal discomfort[4]. A small proportion of COVID-19 patients progress to severe illness requiring hospitalization or intensive care management; among these individuals, mortality owing to Acute Respiratory Distress Syndrome (ARDS) is higher[5]. The estimated average time from symptom-onset to resolution can range from three days to more than three weeks, with a high degree of variability[6]. The COVID-19 public health crisis demands a data science-driven and temporal pathophysiology-informed precision medicine approach for its effective clinical management.

Here we introduce a platform for the augmented curation of the full-spectrum of patient phenotypes from 15,775,993 clinical notes of the Mayo Clinic EHRs for 30,494 patients with confirmed positive/negative COVID-19 diagnosis by PCR testing (see **Methods**). The platform utilizes state-of-the-art transformer neural networks on the unstructured clinical notes to automate entity recognition (e.g. diseases, drugs, phenotypes), quantify the strength of contextual associations between entities, and characterize the nature of association into positive, negative, or other sentiments. We identify specific gastro-intestinal, respiratory, and sensory phenotypes, as well as some of their specific combinations, that appear to be indicative of impending COVID$_{pos}$ diagnosis by PCR testing. This highlights the potential for neural networks-powered EHR curation to facilitate a significantly earlier diagnosis of COVID-19 than currently thought feasible.

## Results

The clinical determination of the COVID-19 status for each patient was conducted using the SARS-CoV-2 PCR (RNA) test approved for human nasopharyngeal and oropharyngeal swab specimens under the U.S. FDA emergency use authorization (EUA)[6]. This PCR test resulted in 14,695 COVID$_{neg}$ patient diagnoses and 635 COVID$_{pos}$ patient diagnoses. In order to investigate the time course of COVID-19 progression in patients, we used BERT-based deep neural networks to extract symptoms and their putative synonyms from the clinical notes for a few weeks prior to, and a few weeks post, the date when the COVID-19 diagnosis test was taken (see **Methods; Table 1**). For the purpose of this analysis, all patients were temporally aligned, by setting the date of COVID-19 PCR testing to 'day 0', and the proportion of patients demonstrating each symptom derived from the EHR over each day of the week preceding and post PCR testing was tabulated (**Table 2**). As a negative control, we included a non-COVID-19 symptom 'dysuria'.

Altered or diminished sense of taste or smell (dysgeusia or anosmia) is the most significantly amplified signal in COVID$_{pos}$ over COVID$_{neg}$ patients in the week preceding PCR testing (**Table 1**; 37.4-fold amplification; p-value = 2.9E-187). This result suggests that anosmia and dysgeusia are likely the most salient early indicators of COVID-19 infection, including in otherwise asymptomatic patients.

In the COVID$_{pos}$ patients, diarrhea is significantly amplified in the week preceding PCR testing (**Table 1**; 2.2-fold; p-value = 3.9E-16). Some of these undiagnosed COVID-19 patients that experience diarrhea may be unintentionally shedding SARS-CoV-2 fecally[7]. Incidentally, epidemiological surveillance by waste water monitoring conducted recently in the state of Massachusetts observed copious SARS-CoV-2 RNA[8]. The amplification of diarrhea in COVID$_{pos}$ over COVID$_{neg}$ patients in the week preceding PCR testing highlights the importance and necessity for washing hands often.

Respiratory difficulty is enriched in the week prior to PCR testing in COVID$_{pos}$ over COVID$_{neg}$ patients (1.9-fold amplification; p-value = 1.1E-22; **Table 1**). Among other common phenotypes with significant enrichments in COVID$_{pos}$ over COVID$_{neg}$ patients, cough has a 1.8-fold amplification (p-value = 9.3E-25), myalgia/arthralgia has a 2.6-fold amplification (p-value = 2E-24), and fever/chills has a 2.1-fold amplification (p-value = 1.3E-36). Rhinitis is also a potential phenotype of COVID$_{pos}$ patients that requires some consideration (1.9-fold amplification, p-value = 1.3E-07). Conjunctivitis, hemoptysis, and respiratory failure have too few COVID$_{pos}$ patients currently for them to be statistically meaningful enrichments, but these phenotypes are worth tracking (**Table 1**). Finally, dysuria was included as a negative control for COVID-19, and consistent with this assumption, 0.94% of COVID$_{pos}$ patients and 0.93% of COVID$_{neg}$ patients had dysuria during the week preceding PCR testing.

Next, we considered the 351 possible pairwise conjunctions of 27 phenotypes for COVID$_{pos}$ versus COVID$_{neg}$ patients in the week prior to the PCR testing date (**Table S1**). As expected from the above results, altered sense of smell or taste (anosmia/dysgeusia) dominates in combination with many of the above symptoms as the most significant combinatorial signature of impending COVID$_{pos}$ diagnosis (particularly along with cough, respiratory difficulty, fever/chills). Examining the other 325 possible pairwise symptom combinations, excluding the altered sense of smell of taste, reveals other interesting combinatorial signals. The combination of cough and diarrhea is noted to be significant in COVID$_{pos}$ over COVID$_{neg}$ patients during the week preceding PCR testing; i.e. cough and diarrhea co-occur in 12.4% of COVID$_{pos}$ patients and in 3.9% of COVID$_{neg}$ patients, indicating a 3.2-fold amplification of this specific symptom combination (BH corrected p-value = 4.0E-17, **Table S1**).

We further investigated the temporal evolution of the proportion of patients with each symptom over the week prior to PCR testing (**Table 2**). Altered sense of taste or smell, cough, diarrhea, fever/chills, and respiratory difficulty were found to be significant discriminators of COVID$_{pos}$ from COVID$_{neg}$ patients between 4 to 7 days prior to PCR testing. During that time period, cough is amplified in the COVID$_{pos}$ patient cohort over the COVID$_{neg}$ patient cohort with an amplification of 3.9-fold (day -7, p-value = 1.4E-09), 4.33-fold (day -6, p-value = 1.8E-12), 2.6-fold (day -5, p-value = 3.9E-04), and 2.2-fold (day -4, p-value = 8.4E-03). The intriguing diminishing odds of cough as a symptom from 7 to 4 days preceding the PCR testing date, suggests this may be a notable temporal pattern. Likewise, diarrhea is amplified in the COVID$_{pos}$ patient cohort over the COVID$_{neg}$ patient cohort with an amplification of 3.2-fold (day -7, p-value = 4.1E-04), 3.0-fold (day -6, p-value = 6.3E-04), 1.9-fold (day -5, p-value = 1.0E-01), and 1.2-fold (day -4, p-value = 6.8E-01). Similarly, fever/chills is also enriched in the COVID$_{pos}$ patient cohort over the COVID$_{neg}$ patient cohort with 5.1-fold (day -7, p-value = 1.3E-12), 5.1-fold (day -6, p-value = 7.3E-15), 3.1-fold (day -5, p-value = 2.2E-05), and 2.6-fold (day -4, p-value = 5.7E-04) amplifications. Respiratory difficulty is likewise enriched in day -7 (3.7-fold, p-value = 4E-07), day -6 (3.8-fold, p-value = 2.9E-08), and day -5 (2.7-fold, p-value = 4.4E-04) among the COVID$_{pos}$ patient cohort. Temporally subsequent to the enriched odds of diarrhea, cough, fever/chills, and respiratory difficulty, we find that change in appetite/intake is amplified in the COVID$_{pos}$ cohort over the COVID$_{neg}$ cohort on day -4 (6.3-fold, p-value = 2.2E-08), day -3 (6.0-fold, p-value = 5.7E-08), and day -2 (4.6-fold, p-value = 5.9E-06). These observations characterize the temporal evolution of specific phenotypes that are enriched in COVID$_{pos}$ patients, preceding and succeeding their PCR testing date.

While explicit identification of SARS-CoV-2 in patients prior to the PCR testing date was not conducted, such prospective validation of our augmented EHR curation approach is being initiated. Nevertheless, this high-resolution temporal overview of the EHR-derived clinical phenotypes as they relate to the SARS-CoV-2 PCR diagnostic testing date for 30,494 patients has revealed specific enriched signals of impending COVID-19 onset. These clinical insights can help modulate social distancing measures and appropriate clinical care for individuals exhibiting the specific gastro-intestinal (diarrhea, change in appetite/intake), sensory (anosmia, dysgeusia) and respiratory phenotypes identified herewith, including for patients awaiting conclusive COVID-19 diagnostic testing results (e.g. by SARS-CoV-2 RNA RT-PCR).

## Discussion

In order to identify potential cells and tissue types that may be associated with the EHR-derived clinical phenotypes observed above for COVID-19 patients, we analyzed Single Cell RNA-seq data using the nferX platform (see **Methods**)[9]. Given recent studies implicating the necessity of both ACE2 and TMPRSS2 for the SARS-CoV-2 lifecycle[10], we scouted for human cells that co-express both genes. This co-expression analysis revealed that specific cell types from the small intestine/colon, nasal cavity, respiratory system, pancreas, urinary tract, and gallbladder co-express both ACE2 and TMPRSS2 (**Figure 1, Figure S1**). Notably, multiple small intestine cell types co-express the two genes. These cell types include enterocytes, enteroendocrine cells, stem cells, goblet cells, and Paneth cells. In the pancreas, the cell types included ductal cells and acinar cells. The kidney cells co-expressing TMPRSS2 and ACE2 include proximal tubular cells, pelvic epithelial cells and type A intercalated cells. Co-expression of TMPRSS2 and ACE2 is also observed in the epithelial cells of the olfactory nasal cavity and the respiratory tract as well as in type

II pneumocytes (albeit at comparatively lower level). While the identified tissues showing co-expression of ACE2 and TMPRSS2 in the gastro-intestinal, respiratory, and sensory systems correlate with the clinical phenotypes of early COVID-19 infection as described above, these insights are conceivably from normal/healthy tissues. This highlight the need for meticulous bio-banking of COVID-19 patient-derived biospecimen and their characterization via single cell RNA-seq and other molecular technologies.

Primary prevention is the most effective method to minimize spread of contagious infectious viruses such as SARS-CoV-2 (**Figure S2**). In addition to population-based strategies such as social distancing, there are significant ongoing efforts to develop a prophylactic solution (**Table S2**). As the immunodominant humoral immune response in patients is directed against the SARS-CoV2 spike protein, many vaccines under investigation target this viral protein. It remains to be determined whether anti-spike protein antibodies induced by natural infection or by vaccines induce neutralizing antibody responses. Chloroquine and its analogues have been shown to inhibit virus replication in-vitro[28]. Whether Chloroquine or Hydroxychloroquine have meaningful effects of SARS-CoV2 replication in patients remains to be understood, and are the subject of clinical trials, both as post-exposure prophylaxis and as treatment (**Table S2**). Hydroxychloroquine was approved by FDA for emergency use in hospitalized COVID-19 patients who are not eligible for clinical trials on April 7, 2020 based on limited clinical data, but concerns have been raised about toxicity and risk of sudden death[29].

Our findings from the EHR analysis of COVID-19 progression can aid in a human pathophysiology enabled summary of the experimental therapies being investigated for COVID-19 (**Figure 2, Table S2**). Some of the earliest phases of intervention attempt to inhibit the entry/replication of SARS-CoV-2 by modulating critical host targets (e.g. renin angiotensin aldosterone system/RAAS inhibitors, ACE2 analogs, serine protease inhibitors) or directly inhibiting the function of viral proteins (e.g. viral RNA-dependent RNA polymerase inhibitors, protease inhibitors, convalescent plasma, synthetic immunoglobulins) (**Box 1, Table S3**). In patients with more advanced stages of disease progression, who suffer from respiratory abnormalities, therapeutics are being advanced to target the inflammatory response that can lead to Acute Respiratory Disease Syndrome (ARDS) and is associated with high mortality (**Box 1**). These include anti-GM-CSF agents, anti-IL-6 agents, JAK inhibitors, and complement inhibitors. Another emerging option for patients at this stage is convalescent plasma, which has shown some clinical benefits in cases of COVID-19 and related viral diseases (SARS-1, MERS) at various stages of severity (**Box 1**). Administration of convalescent plasma containing active specific antiviral antibodies may prevent or attenuate progression to severe disease. Expanded access to convalescent plasma for treatment of patients with COVID-19 has been approved by the FDA for emergency IND use and is available through a nationwide program led by Mayo Clinic (**Box 1**).

In those who become symptomatic, it is imperative that diagnostic testing is done, at dedicated testing sites if available, to confirm diagnosis (**Figure S2**). Meanwhile, patients are recommended to self-quarantine at home, use mask protection when social distancing cannot be obtained, and continue supportive measures. For patients with mild symptoms, such measures may be sufficient given the self-limited nature of viral syndromes. In the event of symptom exacerbation, often marked by worsening respiratory distress, medical evaluation is warranted, and possible hospitalization. The mainstay of treatment for COVID-19, remains supportive care, and as needed supplemental oxygen. Experimental therapies intended to block SARS-CoV2 viral entry and inhibit steps in the viral life cycle necessary for viral replication have been proposed at this early stage (**Figure 2**). The goal of these therapies is to reduce viral load, thus reducing the chance of overwhelming immune reaction by delaying progression of the disease.

Among the proposed treatment options for COVID-19, corticosteroid should be avoided outside a clinical trial, as suggested by the IDSA, until further clinical evidence can be established (www.idsociety.org/practice-guideline/covid-19-guideline-treatment-and-management). This is because there has been conflicting evidence and guidance on steroid use in COVID-19[30]. While steroids can play a role in control of inflammation, a collection of clinical evidence from steroid use in other coronavirus outbreaks suggest that the use of corticosteroids might exacerbate COVID-19-associated lung injury[31].

As patients progress to severe or critical diseases, the primary objective of COVID-19 management is to provide respiratory support and control immune overactivation (**Figure 3, Figure S3**). Patients whose condition deteriorates to critical status primarily decompensate from a respiratory standpoint, but may also develop multi-organ failure (respiratory failure, cardiac failure, renal failure, hypercoagulable state, thrombotic microangiopathy), as well as severe inflammatory responses similar to cytokine release syndrome and eventually reactive hemophagocytosis lymphohistiocytosis syndrome. A major manifestation of respiratory decompensation and cytokine release syndrome is acute respiratory distress syndrome (ARDS). Critical care support such as mechanical support from noninvasive to invasive mechanical ventilation and in, some instances, extracorporeal support, vasopressors, renal replacement therapy, anticoagulation, and are paramount to survival of these critically ill patients per SCC guidelines (SCCM/ESICM 2020). On the other hand, drugs such as immunomodulatory agents often used to treat cytokine release syndrome, may allow for some degree of improvement or recovery either leading into or during severe and critical disease (**Figure 2**).

This study demonstrates how the highly unstructured institutional knowledge can be robustly synthesized using deep learning and neural networks[32]. We started by leveraging a BERT-based deep neural network that was trained on ~18500 sentences containing 250 different phenotypes and symptoms related to cardiovascular, pulmonary, and cardiopulmonary diseases. We found that the model was translatable to the COVID-19 context, achieving an overall 92.7% true positive rate (see **Methods** and **Table S5** for variations across symptoms). While these results speak to the scalability of our deep learning-based approach to teaching machines how to read and comprehend de-identified clinical text, rigorous statistical assessments of neural network performance across disease areas will be essential to further advancing diagnostic applications for clinical care.

Expanding beyond one institution's COVID-19 diagnostic testing and clinical care to the EHR databases of other academic medical centers and health systems will provide a more comprehensive view of clinical phenotypes enriched in $COVID_{pos}$ over $COVID_{neg}$ patients in the days preceding confirmed diagnostic testing. This requires leveraging a privacy-preserving federated software architecture that enables each medical center to retain the span of control of their de-identified EHR databases, while enabling the machine learning models from partners to be deployed in their secure cloud infrastructure. Such seamless multi-institute collaborations over an Augmented Intelligence platform that puts patient privacy and HIPAA-compliance first, is being advanced actively over the Mayo Clinic's Clinical Data Analytics Platform Initiative (CDAP). The capabilities demonstrated in this study for rapidly synthesizing over 8.2 million unstructured clinical notes to develop an EHR-powered clinical diagnosis framework will be further strengthened through such a universal biomedical research platform.

A caveat of relying solely on EHR inference is that mild phenotypes that may not lead to a presentation for clinical care, such as anosmia, may go unreported in otherwise asymptomatic patients. However, the augmented curation approach described here allows for the active monitoring of all such symptoms as they emerge in the EHR; this may accelerate the identification of novel disease symptomatology. As a case in point, the CDC recently expanded its list of possible symptom indicators of COVID-19 to include new loss of taste or smell, headache, muscle pain, and repeated shaking with chills (www.cdc.gov/coronavirus/2019-ncov/symptoms-testing/symptoms.html). As awareness of these symptoms grows, we expect their presence in the EHR to also grow, making statistical enrichment observations, such as those presented here, more robust. Another salient consideration is that as serology-based tests for COVID-19 with high sensitivity and specificity are approved, testing will become more aggressive and "day 0" as defined in the present work will likely occur earlier in the COVID-19 illness course. Additionally, as at-home serology-based testing is advanced, capturing symptoms that precede clinical testing will become increasingly important in order to facilitate the continued development and refinement of disease models; EHR-integrated digital health tools may help address this need. Finally, as multiple COVID-19 testing approaches are pursued and patients begin to receive multiple tests of different types, it will be important to leverage EHR curation tools to assess false positive rates of early serology testing and to gain insight into and optimize test sequencing.

As we continue to understand the diversity of COVID-19 patient outcomes through holistic inference of EHR systems, it is equally important to invest in uncovering the molecular mechanisms and gain cellular/tissue-scale pathology insights through large-scale patient-derived biobanking and multi-omics sequencing. As the anecdotal single cell RNA-seq (scRNA-seq) based co-expression analysis of ACE2 and TMPRSS2 on normal human samples conducted here highlights, the rich heterogeneity of cell types constituting various host tissues can be investigated in great detail by scRNA-seq. To correlate patterns of molecular expression from scRNA-seq with EHR-derived phenotypic signals of COVID-19 disease progression, a large-scale bio-banking system has to be created. Such a system will enable deep molecular insights into COVID-19 to be gleaned and triangulated with SARS-CoV-2 tropism and patient outcomes.

Ultimately, connecting the dots between the temporal dynamics of COVID$_{pos}$ and COVID$_{neg}$ clinical phenotypes across diverse patient populations to the multi-omics signals from patient-derived bio-specimen will help advance a more holistic understanding of COVID-19 pathophysiology. This will set the stage for a precision medicine approach to the diagnostic and therapeutic management of COVID-19 patients.

## BOX 1

**Experimental therapies targeting entry and replication of SARS-CoV-2**

*RAAS inhibitors and ACE2 analogs:* One class of experimental therapies intended to inhibit viral entry and early disease in COVID-19 includes Renin Angiotensin Aldosterone System (RAAS) inhibitors and recombinant ACE2 (**Table S2**). ACE2 is the primary host receptor for SARS-CoV-2, while serine protease TMPRSS2 is implicated in the spike protein priming after viral binding[10]. Recombinant ACE2 has been proposed as an early COVID-19 therapy based on in-vitro data[11]. At this time, the effect of RAAS inhibitors is uncertain in the context of COVID-19. Studies have investigated how ACE expression is modulated by coronavirus infection, and how that relates to lung injury[11]. Trials are ongoing with Angiotensin Receptor Blockers (ARBs) for treatment of COVID-19 by diminishing downstream harmful effects of angiotensin receptor activation (**Figure 2**).

*Serine Protease inhibitors*: Given the TMPRSS2 involvement in viral entry (**Figure 2**), serine protease inhibitors such as Camostat are now under evaluation in trials and should also be considered in the early stages of SARS-CoV-2 infection.

*Viral RNA-dependent RNA polymerase inhibitors*: Of these, Remdesivir, a nucleoside analog, has attracted much attention for in-vivo inhibition of SARS-CoV-2, and a recent observational study of 53 patients who received Remdesivir under compassionate use found that 68% of patients demonstrated improvement in respiratory status after a 10 day regimen[12]. Another nucleoside analog, Galidesivir, is also under evaluation in patients. Yet another viral replication inhibitor in clinical trials is Favipiravir (**Figure 2**). Favipiravir is a broad-spectrum viral RNA dependent RNA polymerase inhibitor that is shown to have in-vivo activity against a wide range of RNA viruses. In one RCT of 240 patients, Favipiravir was found to improve the clinical recovery rate of COVID-19 relative to Umifenovir, a viral entry inhibitor[13] (**Table S2**).

*HIV Protease inhibitors*: This class of medication is widely proposed and used off-label based on postulates that HIV and HCV proteases share structural similarities with those of SARS-CoV-2[14]. Of these, Lopinavir/Ritonavir (combination) has shown promise but was found to have a non-significant benefit in a Randomized Clinical Trial (RCT) of 199 patients in China[15], while Darunavir has shown no significant activity against SARS-CoV-2 in-vitro (**Table S2**)[16]. Multiple randomized, controlled clinical trials are now underway in the USA to determine efficacy of these drugs in the treatment of COVID-19.

*Other Antiviral Agents*: Another emerging option for patients at this stage is convalescent plasma (**Figure 2**), which has shown clinical benefits in cases of COVID-19[17] and related viral diseases (SARS-1, MERS) at various stages of severity[18,19]. Administration of convalescent plasma containing specific antiviral antibodies may prevent or attenuate progression to severe disease. Expanded access to convalescent plasma for treatment of patients with COVID-19 is available through a program led by Mayo Clinic[20]. Synthetic hyperimmune globulins are also under development and evaluation.

**Agents being advanced that target the inflammatory response in COVID-19**

<u>Anti-GM-CSF agents</u> -- A xenograft study found that granulocyte monocyte colony stimulating factor (GM-CSF) neutralization with Lenzilumab significantly reduced production of inflammatory cytokines[21], offering evidence for efficacy of anti-GM-CSF agents in prevention of CART-induced cytokine release syndrome (CRS). Lenzilumab has been approved by the FDA for emergency IND use for CRS in COVID-19, while others such as Mavrilimumab and Gimsilumab aimed at controlling undesired inflammation from myeloid activation will be evaluated in clinical trials.

*Anti-IL-6 agents*: IL-6 is a pro-inflammatory cytokine, regarded as a driver of CRS[22] (**Figure 2A-C**). A recent report suggests IL-6 as a biomarker for respiratory failure in COVID-19[22]. As such, anti-IL-6 agents including Tocilizumab, Sarilumab, and Siltuximab are being evaluated in randomized trials (**Table S2**) and used off-label in severe COVID-19 patients. Tocilizumab was approved for the treatment of CRS in 2017. An observational study of 21 patients with severe COVID-19 pneumonia treated with Tocilizumab showed promising results[23,24].

*Anti-JAK agents*: A number of immunomodulatory agents not linked to CRS are also under trial for COVID-19 (**Figure 2**). Janus kinase (JAK) inhibitors such as Baricitinib, Fedratinib, and Ruxolitinib, indicated for Rheumatoid Arthritis and Myelofibrosis, have been tested in xenograft models for Chimeric Antigen Receptor (CAR) T-cell therapy induced CRS[25]. Ruxolitinib is available under an expanded access program in USA for severely ill COVID-19 patients (**Table S1**) and trials are underway in other countries.

*Anti-Complement agents*: A recent study found that SARS-CoV-2 also binds to MASP2, a key driver of the complement activation pathway, leading to complement hyperactivation in COVID-19 patients[26]. Inhibitors of the terminal complement pathway such as Eculizumab have been tried in individuals with improvements observed after administration in China.

**Agents targeting ventilation/perfusion defects in COVID-19-induced ARDS**

*Vasodilators*: A recent report based on 16 cases in Italy and Germany noted that, contrary to the established understanding in ARDS, COVID-19 patients in ARDS retain relatively high lung compliance[27] and demonstrate ventilation/perfusion defects likely arising from perfusion dysregulation and hypoxic vasoconstriction. Therefore, patients with COVID-19 in ARDS may benefit from vasodilators to address this pathophysiologic mechanism. A trial is underway in China for use of inhaled nitric oxide in patients with mechanical ventilation **(Table S2).**

## Methods

**Augmented curation of EHR patient charts**

The nferX Augmented Curation technology was leveraged to rapidly curate the charts of SARS-CoV-2-positive patients. First, we read through the charts of 100 patients and identified and grouped symptoms into sets of synonymous words and phrases. For example, "SOB", "shortness of breath", and "dyspnea", among others, were grouped into "shortness of breath". We did the same for diseases and medications. For the SARS-CoV2-positive patients, we identified a total of 26 symptom categories (**Table S4**) with 145 synonyms or synonymous phrases. Together, these synonyms and synonymous phrases capture a multitude of ways that symptoms related to COVID-19 are described in the Mayo Clinic Electronic Health Record (EHR) databases.

Next, for charts that had not yet been manually curated, we used state-of-the-art BERT-based neural networks[32] to classify symptoms as being present or not present based on the surrounding phraseology. The neural network used to perform this classification was trained using nearly 250 different phenotypes and 18,500 sentences; it achieves over 96% recall for positive/negative sentiment classification (**Table S6**). We went through individual sentences and either accepted the sentences or rejected and reclassified them. The neural networks were actively re-trained as curation progressed, leading to stepwise increases in curation efficiency and model accuracy. In Step 1 of this process, we labeled 11433 sentences, 8737 of which were labeled as either 'present' or 'not present.' The model trained on this data set (90%-10% training/test split) achieved F1 scores of 0.93 and 0.84 for 'present' and 'not present' classifications, respectively. The model was then applied to an additional 3688 sentences in Step 2, rapidly corrected by a human for classification errors and re-trained to generate a newer version of the model. Step 3 was an iteration of step 2 on an additional 3369 sentences. The model achieved F1 scores of 0.96/0.91 after step 2 and 0.96/0.96 after step 3 for the classification of 'present'/'not present.' Due to the augmented nature of this approach, steps 2 and 3 required successively less input from the human annotator.

This model was applied to 15,775,993 clinical notes across 635 COVID$_{pos}$ patients and 29,859 COVID$_{neg}$ patients. First, the difference between the date on which a particular note was written and the PCR testing date of the patient corresponding to that note formed the relative date measure for that note. The PCR testing date was treated as 'day 0' with notes preceding it assigned 'day-1', 'day-2' and so on. BERT-based neural networks were applied on each note to provide a set of symptoms that were present at that point of time for the patient in question. This map was then inverted to determine for each symptom and relative date the set of unique patients experiencing that symptom.

Because the model had not yet encountered many COVID-19-related symptoms (Table S4), we performed a validation step in which the classifications of 4000 such sentences from the timeframe of highest interest (day 0 to day -7) were manually verified. Sentences arising from templates, such as patient education documentation, accounted for 10.2% of sentences identified. These template sentences were excluded from the analysis. The true positive rate (TPR), defined as the total number of correct classifications by the model divided by the number of total classifications, achieved by the model for classifying all symptoms was 92.7%; the corresponding false positive rate (FPR) was 13.4%. The model achieved true positive rate (TPR) ranging from 91% to 100% for the major symptom categories of Fever/Chills, Cough, Respiratory Difficulty, Headache, Fatigue, Myalgia/Arthralgia, Dysuria, Change in appetite/intake, and Diaphoresis. Classification performance was lower for Altered or diminished sense of taste and smell; here, the true positive rate was 64.4%. Detailed statistics are displayed in **Table S5**.

For each synonymous group of symptoms, we computed the count and proportion of COVID$_{pos}$ and COVID$_{neg}$ patients that were deemed to have that symptom in at least one note between 1 and 7 days prior to their PCR test. We additionally computed the ratio of those proportions which indicates the extent of prevalence of the symptom in the COVID$_{pos}$ cohort as compared to the COVID$_{neg}$ cohort. A standard 2-proportion z hypothesis test was performed, and a p-value was reported for each symptom.

To capture the temporal evolution of symptoms in the COVID$_{pos}$ and COVID$_{neg}$ cohorts, the process described above was repeated considering counts and proportions for each day independently.

Pairwise analysis of phenotypes was performed by considering 351 phenotypic pairs from the original set of 27 individual phenotypes. For each pair, we calculated the number of patients in the COVID$_{pos}$ and COVID$_{neg}$ cohorts wherein both phenotypes occurred at least once in the week preceding PCR testing. With these patient proportions, a Fisher exact test p-value was computed. Benjamini-Hochberg correction was applied to account for multiple hypothesis testing.

This research was conducted under IRB 20-003278, "*Study of COVID-19 patient characteristics with augmented curation of Electronic Health Records (EHR) to inform strategic and operational decisions*". All analysis of EHRs was performed in the privacy-preserving environment secured and controlled by the Mayo Clinic. nference and the Mayo Clinic subscribe to the basic ethical principles underlying the conduct of research involving human subjects as set forth in the Belmont Report and strictly ensures compliance with the Common Rule in the Code of Federal Regulations (45 CFR 46) on Protection of Human Subjects.

**Analysis of cell-types expressing ACE2 and TMPRSS2 using single cell RNAseq**

Since the successful entry of virus in the cell requires priming by cellular host protease – TMPRSS2, we hypothesized that cells that express both TMPRSS2+ and ACE2+ cells could harbor SARS-CoV-2 during the course of infection. Thus, we probed for the expression of ACE2 and TMPRSS2 in all the single-cell studies from human tissues available on the nferX Single Cell platform (https://academia.nferx.com/). For all the tissues that we profiled, we ensured that there is a minimum of 100 cells in the cell population and that there is a minimum of 1% of the cells in the cell population co-expressing (non-zero expression) both TMPRSS2 and ACE2 expression.

# Figure Legends

**Figure 1. Clinical phenotypes of COVID-19 and their connection to single cell RNA-seq co-expression of ACE2-TMPRSS2.** Severity of COVID-19 and associated clinical conditions are shown. Cell types co-expressing SARS-CoV-2 infectivity determinants ACE2 and TMPRSS2 determined by single cell RNA-seq are mapped onto the COVID-19 pathophysiology summary.

**Figure 2. Pathophysiology of COVID-19, associated treatments, and the underlying molecular mechanisms.** While there is no established treatment strategy for COVID-19, several classes of therapeutics have emerged for the medical management of the disease, on the basis of their known mechanisms of action and the pathophysiology of COVID-19.

# Acknowledgments

We thank Murali Aravamudan, Ajit Rajasekharan, Rakesh Barve, Hongfang Liu, and Yanshan Wang for their thoughtful review and feedback on this manuscript. We also thank Andrew Danielsen, Jason Ross, Jeff Anderson, Ahmed Hadad, and Sankar Ardhanari for their support that enabled the rapid completion of this study.

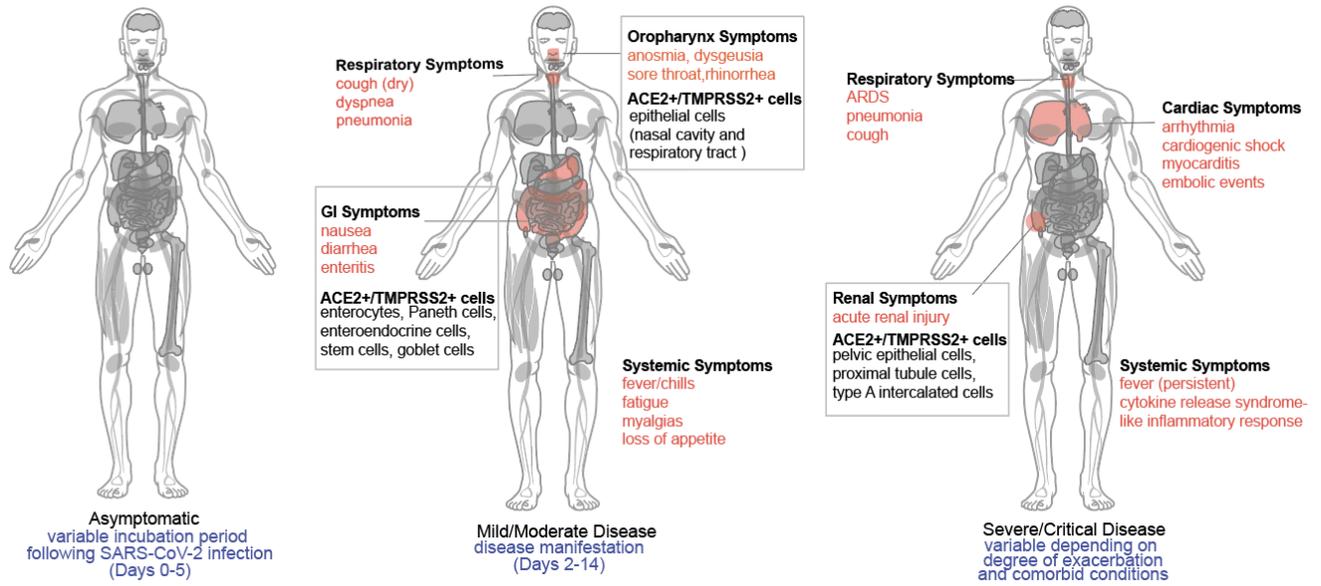

**Figure 1**

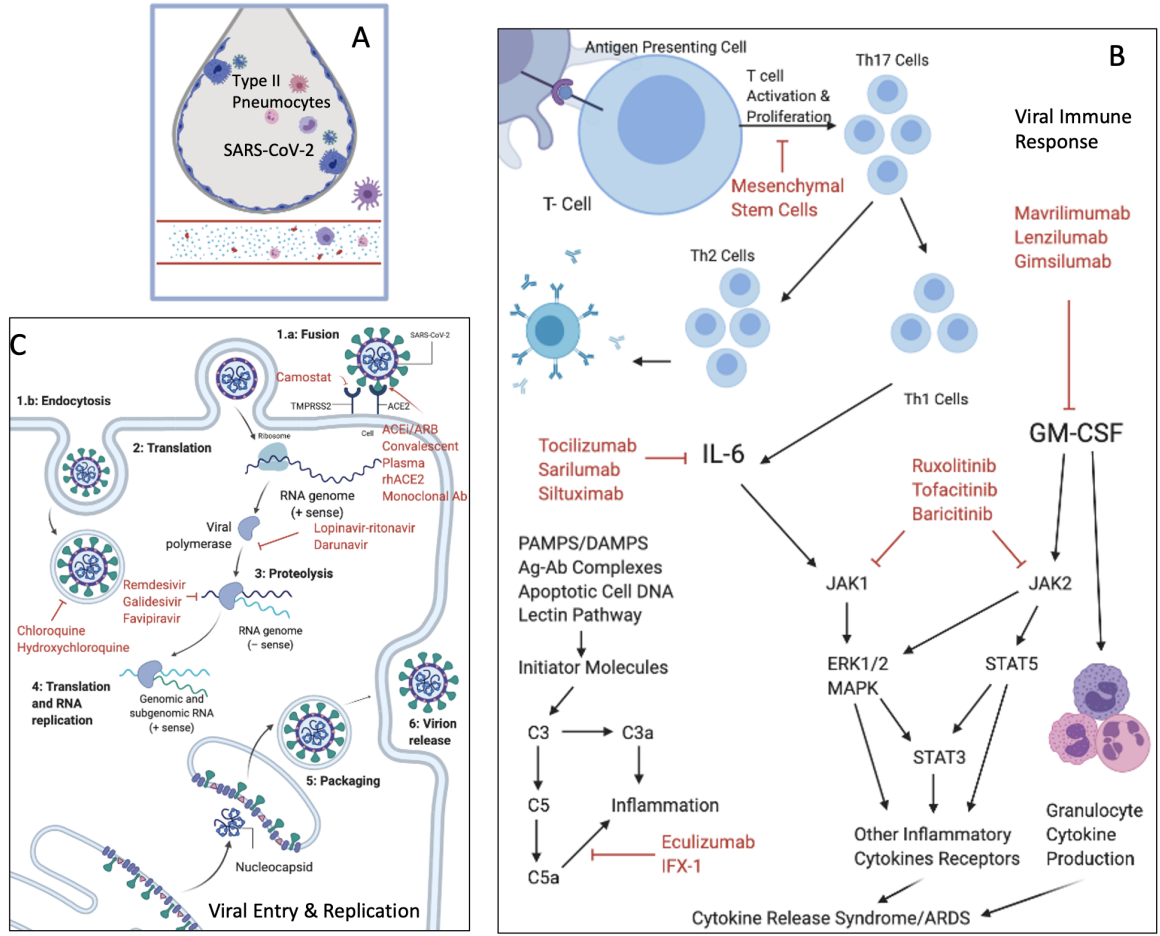

Figure 2

**Table 1. Augmented curation of the unstructured clinical notes from the EHR reveals specific clinically confirmed phenotypes that are amplified in COVID$_{pos}$ patients over COVID$_{neg}$ patients in the week prior to the SARS-CoV-2 PCR testing date.** The key COVID$_{pos}$ amplified phenotypes in the week preceding PCR testing (i.e. day = -7 to day = -1) are highlighted in gray (*p-value < 1E-10*). The ratio of COVID$_{pos}$ to COVID$_{neg}$ proportions represents the fold change amplification of each phenotype in the COVID$_{pos}$ patient set (phenotypes are sorted based on this column). Phenotypes highlighted with a superscript (*) are still rare in COVID$_{pos}$ patients at this time, thus mitigating their statistical significance.

| Phenotype (p-value < 1E-10 in gray) | COVID+ count (N=635) | COVID- count (N = 29859) | COVID+ proportion (N=635) | COVID- proportion (N=29859) | (COVID+/COVID-) relative ratio | 2-tailed p-value |
|---|---|---|---|---|---|---|
| **Altered or diminished sense of taste or smell** | 43 | 54 | 0.07 | 0.00 | 37.44 | 2.95E-187 |
| Conjunctivitis* | 5 | 68 | 0.01 | 0.00 | 3.46 | 4.29E-03 |
| **Myalgia/Arthralgia** | 105 | 1906 | 0.17 | 0.06 | 2.59 | 1.99E-24 |
| **Diarrhea** | 101 | 2183 | 0.16 | 0.07 | 2.18 | 3.91E-16 |
| **Fever / chills** | 230 | 5077 | 0.36 | 0.17 | 2.13 | 1.29E-36 |
| **Respiratory difficulty** | 191 | 4680 | 0.30 | 0.16 | 1.92 | 1.08E-22 |
| Rhinitis | 65 | 1614 | 0.10 | 0.05 | 1.89 | 1.28E-07 |
| **Cough** | 236 | 6105 | 0.37 | 0.20 | 1.82 | 9.36E-25 |
| Cardiac | 33 | 873 | 0.05 | 0.03 | 1.78 | 8.42E-04 |
| Respiratory Failure | 26 | 717 | 0.04 | 0.02 | 1.71 | 6.17E-03 |
| Hemoptysis* | 5 | 139 | 0.01 | 0.00 | 1.69 | 2.42E-01 |
| Diaphoresis | 35 | 979 | 0.06 | 0.03 | 1.68 | 1.89E-03 |
| Change in appetite/intake | 35 | 981 | 0.06 | 0.03 | 1.68 | 1.97E-03 |
| Headache | 85 | 2449 | 0.13 | 0.08 | 1.63 | 2.83E-06 |
| Congestion | 73 | 2228 | 0.11 | 0.07 | 1.54 | 1.40E-04 |
| Fatigue | 93 | 3245 | 0.15 | 0.11 | 1.35 | 2.55E-03 |
| Neuro | 50 | 2035 | 0.08 | 0.07 | 1.16 | 2.96E-01 |
| Pharyngitis | 51 | 2129 | 0.08 | 0.07 | 1.13 | 3.83E-01 |
| Dysuria | 6 | 277 | 0.01 | 0.01 | 1.02 | 9.64E-01 |
| Generalized symptoms | 65 | 3502 | 0.10 | 0.12 | 0.87 | 2.47E-01 |
| Chest pain/pressure | 49 | 2738 | 0.08 | 0.09 | 0.84 | 2.09E-01 |
| GI upset | 67 | 3806 | 0.11 | 0.13 | 0.83 | 1.00E-01 |
| Dermatitis* | 13 | 895 | 0.02 | 0.03 | 0.68 | 1.63E-01 |
| Otitis* | 5 | 435 | 0.01 | 0.01 | 0.54 | 1.62E-01 |
| Wheezing | 19 | 1917 | 0.03 | 0.06 | 0.47 | 4.56E-04 |
| Dry mouth | 0 | 94 | 0.00 | 0.00 | 0.00 | 1.57E-01 |

**Table 2. Temporal analysis of the EHR clinical notes for the week preceding PCR testing (i.e. day -7 to day -1), leading up to the day of PCR testing (day 0) in COVID$_{pos}$ and COVID$_{neg}$ patients.** Temporal enrichment for each symptom is quantified using the ratio of COVID$_{pos}$ patient proportion over the COVID$_{neg}$ patient proportion for each day. The patient proportions in rows labeled 'Positive (n = 635)' and 'Negative (n = 29859)' are represented as percentages.

| Phenotype | COVID-19 PCR (N = 30494) | Day = -7 | Day = -6 | Day = -5 | Day = -4 | Day = -3 | Day = -2 | Day = -1 |
|---|---|---|---|---|---|---|---|---|
| **Altered or diminished sense of taste or smell** | Positive (n = 635) | 0.47% | 0.31% | 0.31% | 0.16% | 0.16% | 0.00% | 0.63% |
| | Negative (n = 29859) | 0.01% | 0.00% | 0.01% | 0.01% | 0.01% | 0.03% | 0.08% |
| | Ratio (Positive/Negative) | 70.53 | - | 23.51 | 15.67 | 15.67 | 0.00 | 7.84 |
| | p-value | 1.19E-19 | 3.08E-22 | 8.26E-08 | 1.33E-03 | 1.33E-03 | 6.45E-01 | 6.06E-06 |
| **Cough** | Positive (n = 635) | 2.83% | 3.31% | 2.05% | 1.73% | 1.73% | 0.79% | 12.60% |
| | Negative (n = 29859) | 0.72% | 0.76% | 0.78% | 0.79% | 0.90% | 1.19% | 13.03% |
| | Ratio (Positive/Negative) | 3.94 | 4.33 | 2.63 | 2.20 | 1.92 | 0.66 | 0.97 |
| | p-value | 1.40E-09 | 1.82E-12 | 3.88E-04 | 8.42E-03 | 3.07E-02 | 3.54E-01 | 7.50E-01 |
| **Diarrhea** | Positive (n = 635) | 1.42% | 1.42% | 0.94% | 0.63% | 0.63% | 0.16% | 5.20% |
| | Negative (n = 29859) | 0.45% | 0.47% | 0.49% | 0.51% | 0.57% | 0.67% | 3.99% |
| | Ratio (Positive/Negative) | 3.16 | 3.04 | 1.95 | 1.23 | 1.10 | 0.24 | 1.30 |
| | p-value | 4.08E-04 | 6.38E-04 | 1.03E-01 | 6.82E-01 | 8.50E-01 | 1.14E-01 | 1.24E-01 |
| **Fever / chills** | Positive (n = 635) | 2.68% | 3.15% | 2.05% | 1.89% | 1.57% | 0.63% | 11.65% |
| | Negative (n = 29859) | 0.53% | 0.61% | 0.65% | 0.71% | 0.73% | 1.03% | 10.54% |
| | Ratio (Positive/Negative) | 5.06 | 5.14 | 3.15 | 2.66 | 2.17 | 0.61 | 1.11 |
| | p-value | 1.33E-12 | 7.33E-15 | 2.20E-05 | 5.71E-04 | 1.39E-02 | 3.23E-01 | 3.66E-01 |
| **Respiratory Difficulty** | Positive (n = 635) | 2.20% | 2.52% | 1.89% | 1.89% | 1.10% | 0.63% | 7.87% |
| | Negative (n = 29859) | 0.60% | 0.66% | 0.70% | 0.68% | 0.82% | 1.07% | 9.43% |
| | Ratio (Positive/Negative) | 3.70 | 3.80 | 2.71 | 2.78 | 1.35 | 0.59 | 0.83 |
| | p-value | 3.96E-07 | 2.93E-08 | 4.39E-04 | 3.11E-04 | 4.31E-01 | 2.89E-01 | 1.83E-01 |
| **Change in appetite/intake** | Positive (n = 635) | 0.16% | 0.16% | 0.16% | 1.26% | 1.26% | 1.26% | 2.36% |
| | Negative (n = 29859) | 0.15% | 0.16% | 0.18% | 0.20% | 0.21% | 0.27% | 1.45% |
| | Ratio (Positive/Negative) | 1.02 | 0.96 | 0.89 | 6.27 | 5.97 | 4.59 | 1.63 |
| | p-value | 9.83E-01 | 9.67E-01 | 9.05E-01 | 2.17E-08 | 5.75E-08 | 5.93E-06 | 5.75E-02 |
| **Respiratory failure** | Positive (n = 635) | 0.63% | 0.47% | 0.47% | 0.47% | 0.16% | 0.31% | 1.26% |
| | Negative (n = 29859) | 0.28% | 0.27% | 0.30% | 0.30% | 0.29% | 0.33% | 1.02% |
| | Ratio (Positive/Negative) | 2.27 | 1.76 | 1.59 | 1.55 | 0.54 | 0.96 | 1.23 |
| | p-value | 9.99E-02 | 3.28E-01 | 4.28E-01 | 4.51E-01 | 5.34E-01 | 9.54E-01 | 5.62E-01 |
| **Headache** | Positive (n = 635) | 0.79% | 1.26% | 1.26% | 0.79% | 0.47% | 0.16% | 4.25% |
| | Negative (n = 29859) | 0.39% | 0.48% | 0.49% | 0.41% | 0.45% | 0.54% | 4.13% |
| | Ratio (Positive/Negative) | 2.03 | 2.65 | 2.58 | 1.93 | 1.06 | 0.29 | 1.03 |
| | p-value | 1.14E-01 | 5.19E-03 | 6.69E-03 | 1.42E-01 | 9.19E-01 | 1.93E-01 | 8.75E-01 |
| **Fatigue** | Positive (n = 635) | 1.10% | 0.94% | 0.94% | 0.63% | 0.47% | 0.31% | 5.51% |
| | Negative (n = 29859) | 0.65% | 0.73% | 0.75% | 0.73% | 0.71% | 1.00% | 5.25% |
| | Ratio (Positive/Negative) | 1.69 | 1.29 | 1.27 | 0.87 | 0.66 | 0.31 | 1.05 |
| | p-value | 1.67E-01 | 5.31E-01 | 5.67E-01 | 7.76E-01 | 4.74E-01 | 8.24E-02 | 7.68E-01 |
| **Myalgia/Arthralgia** | Positive (n = 635) | 0.94% | 1.57% | 1.26% | 0.94% | 0.47% | 0.16% | 5.67% |
| | Negative (n = 29859) | 0.19% | 0.20% | 0.26% | 0.23% | 0.20% | 0.33% | 3.37% |
| | Ratio (Positive/Negative) | 4.95 | 7.84 | 4.82 | 4.03 | 2.31 | 0.47 | 1.68 |
| | p-value | 3.46E-05 | 8.16E-13 | 2.66E-06 | 3.81E-04 | 1.44E-01 | 4.41E-01 | 1.56E-03 |
| **Congestion** | Positive (n = 635) | 0.79% | 0.79% | 0.94% | 0.47% | 0.47% | 0.00% | 2.68% |
| | Negative (n = 29859) | 0.27% | 0.30% | 0.28% | 0.28% | 0.26% | 0.38% | 3.79% |
| | Ratio (Positive/Negative) | 2.87 | 2.58 | 3.40 | 1.66 | 1.79 | 0.00 | 0.71 |
| | p-value | 1.65E-02 | 3.17E-02 | 2.05E-03 | 3.83E-01 | 3.17E-01 | 1.19E-01 | 1.44E-01 |
| **Rhinitis** | Positive (n = 635) | 0.79% | 0.79% | 0.63% | 0.31% | 0.16% | 0.00% | 2.36% |
| | Negative (n = 29859) | 0.15% | 0.19% | 0.18% | 0.16% | 0.20% | 0.26% | 2.89% |
| | Ratio (Positive/Negative) | 5.34 | 4.05 | 3.48 | 1.96 | 0.78 | 0.00 | 0.82 |
| | p-value | 6.76E-05 | 1.12E-03 | 1.02E-02 | 3.42E-01 | 8.08E-01 | 1.94E-01 | 4.34E-01 |
| **Dysuria** | Positive (n = 635) | 0.00% | 0.00% | 0.00% | 0.00% | 0.16% | 0.00% | 0.47% |
| | Negative (n = 29859) | 0.04% | 0.05% | 0.04% | 0.04% | 0.05% | 0.07% | 0.41% |
| | Ratio (Positive/Negative) | 0.00 | 0.00 | 0.00 | 0.00 | 2.94 | 0.00 | 1.16 |
| | p-value | 5.99E-01 | 5.60E-01 | 6.13E-01 | 6.29E-01 | 2.72E-01 | 4.94E-01 | 8.03E-01 |

# SUPPLEMENTARY MATERIAL

**Figure S1. Cell-types connected to pathophysiology of COVID-19 as inferred from high expression of ACE2 and TMPRSS2 in human scRNA seq datasets.** A scatter plot depicting the expression of ACE2 and TMPRSS2 inferred from the single-cell RNA-seq profiling of human tissues using nferX single cell platform. The x-axis represents the mean ln(cp10k+1) expression of ACE2 in all the cells and the y-axis represents the mean ln(cp10k+1) expression of TMPRSS2 in the corresponding cell-types from respective tissues. The colors on the scatter plot depicts the tissue origins. The size of the points on the scatter plot represents the percentage of single cells in the cell-type that co-express ACE2 and TMPRSS2 (non-zero expression).

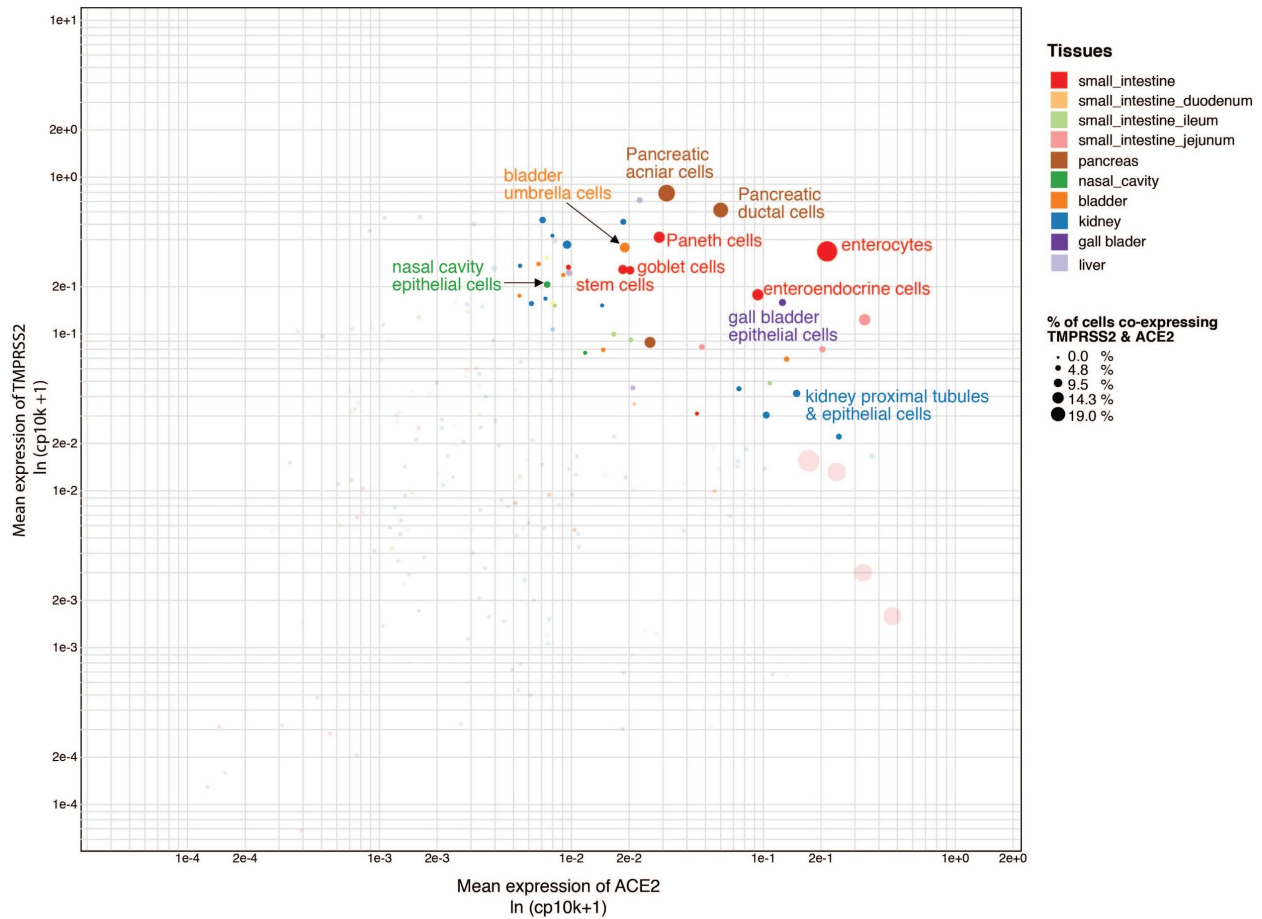

**Figure S2. Disease progression of COVID-19 can be divided into multiple stages, and appropriate therapeutics can be chosen based on the specific pathophysiological mechanisms.** Using nferX Knowledge Synthesis, the most associated molecular markers at each step of disease progression are also identified (see *Supplementary Methods* for details on nferX knowledge synthesis). In order to capture biomedical literature based associations, the nferX platform defines two scores: a "local score" and a "global score", as described previously (Park, J. et al. Recapitulation and Retrospective Prediction of Biomedical Associations Using Temporally-enabled Word Embeddings. doi:10.1101/627513).

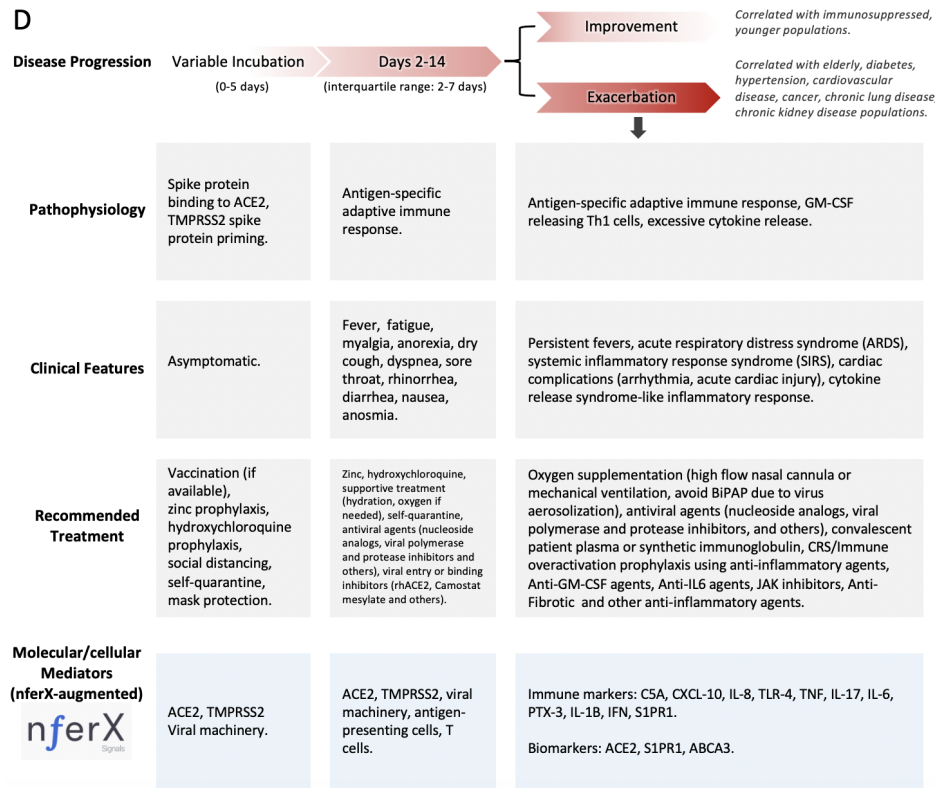

**Figure S3. nferX-derived associations of COVID-19 treatment options to clinical phenotypes. (A)** Schematic of the derivation of nferX local and global scores quantifying associations between concepts from across the literature. **(B)** Heatmap of nferX Local Scores capturing associations discussed in the literature between select COVID-19 treatment drugs and COVID-19 related phenotypes. In order to capture biomedical literature-based associations, the nferX platform defines two scores: a "local score" and a "global score", as described previously (Park, J. et al. Recapitulation and Retrospective Prediction of Biomedical Associations Using Temporally-enabled Word Embeddings. doi:10.1101/627513).

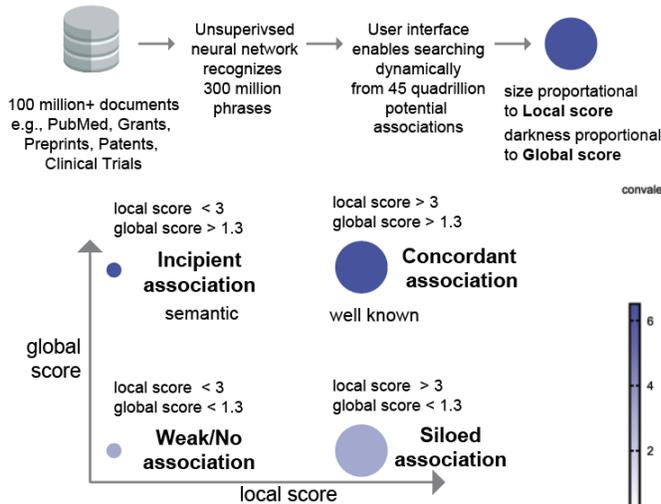
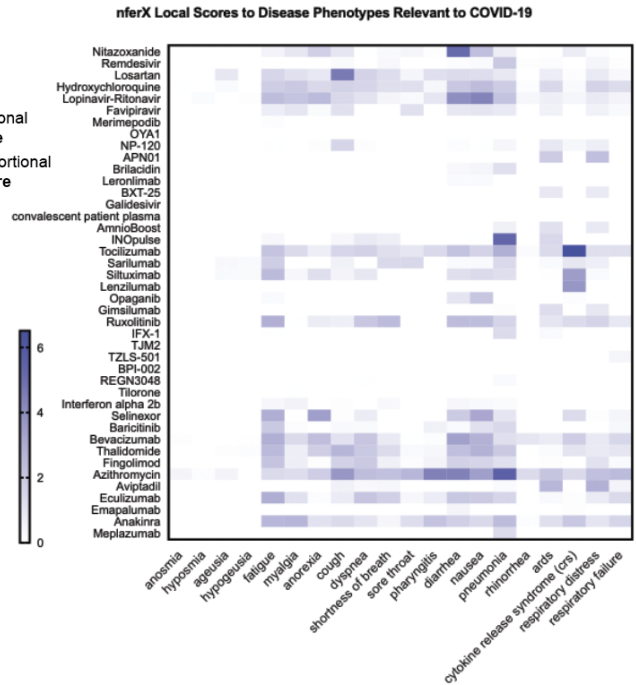

**Table S1. Pairwise analysis of phenotypes in the COVID$_{pos}$ and COVID$_{neg}$ cohorts.** The pairwise symptom combinations with p-value < 1E-06 are summarized.

| Phenotype 1 | Phenotype 2 | COVID+ count (N=635) -- Only > 30 shown | COVID- count (N=29859) | COVID+ % (N=635) | COVID- % (N=29859) | (COVID+)/(COVID-) ratio -- Only > 2.5 shown | raw p-value -- Only < 1E-06 | BH-corrected p-value |
|---|---|---|---|---|---|---|---|---|
| Altered or diminished sense of taste or smell | Cough | 38 | 31 | 5.98 | 0.10 | 57.64 | 9.22E-46 | 2.55E-43 |
| Fever / chills | Altered or diminished sense of taste or smell | 37 | 32 | 5.83 | 0.11 | 54.37 | 5.47E-44 | 7.54E-42 |
| Fever / chills | Respiratory difficulty | 173 | 2558 | 27.24 | 8.57 | 3.18 | 7.46E-42 | 6.86E-40 |
| Fever / chills | Cough | 204 | 3490 | 32.13 | 11.69 | 2.75 | 3.35E-41 | 2.31E-39 |
| Altered or diminished sense of taste or smell | Respiratory difficulty | 34 | 29 | 5.35 | 0.10 | 55.13 | 1.22E-40 | 6.72E-39 |
| Fever / chills | Myalgia/Arthralgia | 86 | 1300 | 13.54 | 4.35 | 3.11 | 1.25E-19 | 3.45E-18 |
| Diarrhea | Cough | 79 | 1175 | 12.44 | 3.94 | 3.16 | 1.89E-18 | 4.00E-17 |
| Cough | Myalgia/Arthralgia | 82 | 1293 | 12.91 | 4.33 | 2.98 | 1.02E-17 | 2.02E-16 |
| Fever / chills | Diarrhea | 78 | 1238 | 12.28 | 4.15 | 2.96 | 9.87E-17 | 1.70E-15 |
| Diarrhea | Myalgia/Arthralgia | 43 | 502 | 6.77 | 1.68 | 4.03 | 1.05E-13 | 1.52E-12 |
| Respiratory difficulty | Myalgia/Arthralgia | 61 | 993 | 9.61 | 3.33 | 2.89 | 7.92E-13 | 1.04E-11 |
| Fever / chills | Headache | 76 | 1415 | 11.97 | 4.74 | 2.53 | 9.99E-13 | 1.25E-11 |
| Diarrhea | Respiratory difficulty | 59 | 1008 | 9.29 | 3.38 | 2.75 | 1.23E-11 | 1.48E-10 |
| Fever / chills | Rhinitis | 57 | 985 | 8.98 | 3.30 | 2.72 | 4.25E-11 | 4.88E-10 |
| Respiratory difficulty | Rhinitis | 50 | 799 | 7.87 | 2.68 | 2.94 | 5.59E-11 | 6.17E-10 |
| Respiratory difficulty | Headache | 62 | 1147 | 9.76 | 3.84 | 2.54 | 8.51E-11 | 9.03E-10 |
| Diarrhea | Headache | 37 | 602 | 5.83 | 2.02 | 2.89 | 2.92E-08 | 2.60E-07 |
| Rhinitis | Headache | 31 | 475 | 4.88 | 1.59 | 3.07 | 1.14E-07 | 9.82E-07 |
| Myalgia/Arthralgia | Headache | 41 | 760 | 6.46 | 2.55 | 2.54 | 1.91E-07 | 1.50E-06 |

**Table S2. Collection of treatments undergoing clinical trials for COVID-19.**

| Drug Class | Drug | Proposed Mechanism of Action | NCT Number |
|---|---|---|---|
| **Renin-Angiotensin-Aldosterone System (RAAS) Inhibitors and recombinant ACE2** | ACE inhibitors | Inhibits viral entry and downstream activation of angiotensin mediated inflammation. Effects of ACE inhibition under evaluation | NCT04330300, NCT04322786 NCT04330300 |
| | Angiotensin Receptor Blockers | Inhibits angiotensin receptor. Given that SARS-CoV-2 uses ACE2 in viral entry, ARBs are under evaluation for potential efficacy and/or risk in COVID. | NCT04335123, NCT04328012, and more |
| | Recombinant human ACE2 (rhACE2) | SARS-CoV infection reduces ACE2 expression and that administration of recombinant ACE2 following infection can lead to reversal of lung injury | NCT04287686, NCT04335136 |
| **Antiviral Agents** | Remdesivir | Nucleoside analog reported to improve respiratory status in 70% of patients (of 53 patients total) in an observation study. RCTs underway. | NCT04292899, NCT04292730 and more |
| | Favipiravir | Viral RNA polymerase inhibitor. RCT showed Favipiravir improved the clinical recovery rate of COVID relative to Umifenovir, a viral entry inhibitor | NCT04336904, NCT04333589 and more |
| | Lopinavir/ritonavir | Protease inhibitor used in HIV infection. Found to have a non-significant benefit in an RCT of 199 patients in China. Larger clinical studies in progress. | NCT04295551, NCT04331470 and more |
| | Umifenovir | Broad-spectrum antiviral that inhibits membrane fusion between virus and target hose. Initially developed for influenza. | NCT04260594, NCT04273763 |
| | Ribavirin | Nucleoside analog with broad spectrum activity against RNA and DNA virus. | NCT04276688 |
| | Emtricitabine/tenofovir | Nucleoside reverse-transcriptase inhibitor used for HIV infection being evaluated for COVID prevention in healthcare workers | NCT04334928 |
| | Oseltamivir | sialidase used in influenza. Proposed for role of sialic acid in viral entry of coronavirus | NCT02735707, NCT04261270 |
| | DAS181 | inhaled sialidase, initially developed for parainfluenza infection, proposed given the role of sialic acid in the viral entry of coronavirus | NCT04324489, NCT04298060 and more |
| | Darunavir | Protease inhibitor used in HIV infection, proposed for potential antiviral efficacy. | NCT04252274 |
| | ASC09 | a novel protease inhibitor similar in structure to darunavir. Initially developed for HIV infection. | NCT04261907 |
| | Meplazumab | Shown to bind to spike protein of SARS-CoV-2 and found to reduce severity of COVID in RCT of 17 patients in China [1] | NCT04275245 |
| | Interferon Beta-1A | Shown to have in-vitro inhibitory activity against SARS-CoV2[2] | NCT04315948, NCT02735707 and more |
| | Interferon Beta-1B | | |
| | Levamisole + Budesonide + Formoterol inhaler | Found to bind to Papaine Like Protease, which is implicated in SARS-CoV-2 virulence | NCT04331470 |
| | Camostat Mesilate | Serine protease proposed for in-vitro inhibition of MERS-CoV entry[3] | NCT04321096, NCT04338906 |
| | Convalescent plasma | Antibodies in recovered patients. Synthetic antibodies under development | NCT04264858 |
| **Anti-IL6 Agents** | Tocilizumab | IL-6 inhibitor proposed to reduce risk of cytokine release syndrome in COVID | NCT04317092, NCT04320615, and more |
| | Siltuximab | IL-6 inhibitor proposed to reduce risk of cytokine release syndrome in COVID | NCT04330638, NCT04329650 |
| | Sarilumab | IL-6R inhibitor proposed to reduce risk of cytokine release syndrome in COVID | NCT04315298, NCT04324073, and more |
| **Complement Inhibitors** | Eculizumab | proposed for anti-C5a activity to inhibit complement overactivation in COVID | NCT04288713 |
| | IFX-1 | | NCT04333420 |
| **JAK Inhibitors** | Baricitinib | JAK inhibitor proposed for anti-inflammatory activity to attenuate immune overactivation in severe COVID | NCT04340232, NCT04320277 and more |
| | Tofacitinib | | NCT04332042 |
| | Ruxolitinib | | NCT04337359, NCT04331665 and more |
| **Anti-GM-CSF Agents** | Mavrilimumab | anti-GM-CSF agent proposed for anti-inflammatory effects in the event of COVID-induced cytokine release syndrome | NCT04337216 |
| | Lenzilumab | anti-GM-CSF agent. Approved under emergency IND use. | |
| | Gimsilumab | anti-GM-CSF agent about to begin clinical trials. | |
| **Other Anti-Inflammatory or Anti-Fibrotic Agents** | Piclidenoson | initially developed for rheumatoid arthritis. Binds to A3AR, which may attenuate CRS. Also shown to have antiviral activity against RNA virus[4] | NCT04333472 |
| | Anakinra | IL-1a antagonist initially developed for rheumatoid arthritis. | NCT04330638, NCT04339712 |
| | Emapalumab | Anti-IFN gamma agent proposed for anti-inflammatory effect in combination with anakinra | NCT04324021 |
| | Thalidomide | anti-TNF agent used in multiple myeloma and graft-versus-host diseases proposed for immunomodulatory effects | NCT04273581 |
| | Colchicine | Proposed for inhibition of inflammation caused by NLRP3 inflammasome, through inhibition of tubulin polymerization, and potential effects on cellular adhesion molecules and inflammatory chemokines | NCT04326790, NCT04328480 and more |
| | Fingolimod | Immunomodulatory agent used in multiple sclerosis. | NCT04280588 |
| | Aviptadil | Synthetic VIP peptide proposed for immunomodulatory effects[5] | NCT04311697 |
| | BLD-2660 | Selective CAPN inhibitor initially developed for pulmonary fibrosis. Studies found that CAPN inhibition led to reduce replication of SARS-CoV-1[6] | NCT04334460 |
| | Bevacizumab | anti-VEGF antibody proposed to attenuate increases in vascular permeability in the COVID-induced vascular inflammation | NCT04305106, NCT04275414 |
| | Bromhexine | Increases secretion of mucus components in the respiratory tract and alleviates respiratory inflammation.[7] | NCT04340349, NCT04273763 |
| | CD24Fc | Initially developed for GVHD Shown to reduce inflammation by binding to DAMP and Siglec G/10 to modulate immune response[8] | NCT04317040 |
| | Carrimycin | Under evaluation in China for efficacy against upper respiratory tract diseases | NCT04286503 |
| | Nintedanib | tyrosine kinase inhibitor initially developed for idiopathic pulmonary fibrosis, likely approved for anti-inflammatory and anti-fibrotic activity | NCT04338802 |
| | PUL-042 | TLR 2/6/9 agonist proposed to prevent COVID | NCT04313023, NCT04312997 |
| | Defibrotide | Oligonucleotide initially approved in hepatic veno-occlusive diseases. Likely proposed for effect in modulation of endothelial injury. | NCT04335201 |
| | Ibrutinib | BTK inhibitor used for B-cell malignancies, but also has activity against TEC family kinase. In-vivo evidence for lung protection in viral infection[9] | |
| **Vasodilators or other agents targeting effects of hypoxic vasoconstriction** | Sildenafil | Vasodilator likely proposed to attenuate perfusion dysregulation and hypoxic vasoconstriction in COVID ARDS | NCT04304313 |
| | Inhaled nitric oxide | | NCT04306393 |
| | Sargramostim | inhaled sargramostim, which induces hematopoiesis, proposed to attenuates acute hypoxic respiratory failure in COVID | NCT04326920 |
| **Vaccines** | mRNA-1273 | Clinical trial sponsored by NIAID underway at Kaiser Permanente Washington Health Research Institute (KPWHRI) in Seattle. | NCT04283461 |
| | INO-4800 | Inovio announced an accelerated timeline for the development of the vaccine on 03 March. Trial underway. | NCT04336410 |
| | BCG Vaccine | Suggested to be protective against COVID with limited data[10] | NCT04328441, NCT04327206 |
| **Mesenchymal Stem Cell (MSC) Therapy** | NestCell®, Human Umbilical Derived MSC I, Dental Pulp MSC, Wharton's Jelly-MSCs | MSCs shown to reduce nonproductive inflammation and affect tissue regeneration and is being evaluated in patients with ARDS | NCT04315987 NCT04336254 NCT04313322 And more |
| | MSC + Ruxolitinib | MSC+ ruxolitinib being evaluated in severe COVID-19 | ChiCTR2000029580 |
| | Exosomes allogeneic adipose MSC (MSCs-Exo) | Pilot pilot clinical trial performed to explore the safety and efficiency of aerosol inhalation of the MSCs-Exo in patients with severe COVID-19 | NCT04276987 |

*Table S2 - References*

**Table S3. Disease Progression of COVID-19 and Associated Treatment**

| Diseases Progression | Symptoms and other Clinical Indicators | Recommended Treatment |
|---|---|---|
| Prior to or at time of Exposure | Asymptomatic | ● Vaccination, when available<br>● Zinc Prophylaxis<br>● Hydroxychloroquine Prophylaxis<br>● Social Distancing<br>● Self-Quarantine<br>● Mask Protection |
| Early Infection, without hospitalization | Fever<br>Anosmia/Ageusia<br>Mild respiratory symptoms (cough, dyspnea, sore throat, rhinorrhea)<br>GI symptoms (diarrhea, nausea, abdominal discomfort)<br><br>Focal consolidation on CXR; Focal ground glass opacity on CT | ● Zinc<br>● Hydroxychloroquine<br>● Supportive Treatment (Hydration, Oxygen if needed)<br>● Self-Quarantine<br>● Antiviral agents (nucleoside analogs, viral polymerase and protease inhibitors, and others)<br>● Viral entry or binding inhibitors (rhACE2, Camostat mesylate and others): |
| Moderate diseases, requiring supplemental oxygen | Persistent Fever<br>Moderate respiratory and GI symptoms<br>Dyspnea with increased oxygen requirement<br><br>Oxygen desaturation (<93%)<br>Diffuse Ground Glass Opacity on CT Chest | ● Oxygen supplementation (high flow nasal cannula, avoid BiPAP due to virus aerosolization)<br>● Antiviral agents (nucleoside analogs, viral polymerase and protease inhibitors, and others)<br>● Convalescent patient plasma or Synthetic immunoglobulin<br>● CRS/Immune overactivation prophylaxis using anti-inflammatory agents<br>● Anti-GM-CSF Agents<br>● Anti-IL6 Agents<br>● JAK Inhibitors |
| Severe diseases requiring mechanical ventilation, with immune overactivation | ARDS<br>Cytokine Release Syndrome-like inflammatory response<br>Multi-organ involvement<br><br>Elevation of IL-6, Ferritin & other inflammatory markers<br><br>Diffuse bilateral coalescent opacities | Anti-Inflammatory agents, including<br>● Anti-IL6 Agents<br>● Complement Inhibitors<br>● Anti-GM-CSF Agents<br>● JAK Inhibitors<br>● Anti-Fibrotic Agents |

**Table S4. Symptoms and their synonyms used for the EHR analysis.**

| Symptom/Finding | Synonyms/related entities identified in EHR |
|---|---|
| Fever / chills | fever, fevers, chill, chills, tactile fever, felt warm, subjective fever |
| Altered or diminished sense of taste or smell | change in smell, lost her sense of smell and taste, bitter taste in his mouth, no sense of taste or smell, no sense of smell or taste, decrease in smell, decreased sense of taste, change in her sense of taste and smell, decrease in smell and taste, ageusia, change in taste, lost her sense of taste and smell, dysgeusia, everything smells and tastes terrible, anosmia, altered smell, loss of taste and smell, change in his sense of smell and taste, altered sense of taste and smell, decrease in taste and smell, decreased taste, bitter taste in her mouth, taste is altered, lost his sense of smell, decreased smell, altered taste, altered sense of smell and taste, change in taste and smell, lost his sense of taste and smell, change in her sense of smell and taste, everything tastes and smells terrible, lost his sense of taste and smell, lost her sense of smell, loss of smell and taste, decrease in taste, decrease taste, bitter taste, no smell or taste, no taste or smell, anosmia/dysgeusia, decrease smell, change in smell and taste, change in his sense of taste and smell |
| Diarrhea | loose stools, soft stools, watery diarrhea, soft stool, diarrhea, watery bm, vomiting diarrhea, loose stool, diarrhea vomiting |
| GI upset | abdominal pain, nausea, abdominal cramping, posttussive emesis, stomach ache, emesis, vomiting, nausea vomiting abdominal pain, stomach cramping, vomiting diarrhea, diarrhea vomiting |
| Wheezing | wheezing |
| Respiratory difficulty | increased oxygen demands, lower respiratory symptoms, tachypnea, tachypneic, labored breathing, dyspnea with walking, dyspnea on exertion, shortness of breath, sob with exercise, dyspnea, sob |
| Respiratory failure | Respiratory failure |
| Cough | cough, cough that is nonproductive, productive cough, cough np p, dry cough, cough productive, cough that is productive, nonproductive cough, coughing, non productive cough |
| Hemoptysis | hemoptysis, blood-tinged sputum, red-tinged sputum |
| Chest pain/pressure | chest congestion, chest tightness, pleuritic chest pain, chest heaviness, chest pain, tightness of the chest, chest discomfort |
| Congestion | sinus pressure, congestion, head congestion, stuffy nose, congested, nasal congestion, congestion rhinorrhea, facial pressure, sinus congestion |
| Rhinitis | rhinitis, itchy eyes and nose, runny nose, tickling in nose, sniffles, rhinorrhea, sneezing, congestion rhinorrhea |
| Myalgia/Arthralgia | body ache, myalgia, muscle aches, arthralgias, sore neck, sore muscles, muscle discomfort, joints became sore, body aches, aches and pains, achy joints, arthralgia, myalgias], Generalized symptoms: [weakness, feeling run down, cold, malaise, generalized weakness, influenza like symptoms, weak, feeling poorly |
| Generalized symptoms | cold, generalized weakness, malaise, weakness, weak, influenza like symptoms, feeling run down, activity change |
| Fatigue | fatigued, energy level is diminished, lethargy, fatigue, energy level is poor, activity change, sleeping more than usual, lethargic |
| Diaphoresis | diaphoretic, diaphoresis, night sweats, sweaty, sweating, sweats |
| Pharyngitis | scratchy throat, throat discomfort, tingly throat, throat irritation, sore throat |
| Headache | headache, HA, headaches, HA's, sinus headache |
| Dry mouth | dry mouth, xerostomia |
| Change in appetite/intake | decrease in appetite, diminished appetite, appetite is poor, appetite change, appetite is diminished, appetite has been fluctuating, poor appetite, decreased appetite, anorexia, not eating and drinking, little appetite, no appetite, not eating or drinking |
| Conjunctivitis | itchy eyes and nose, red eyes, watery eyes, pink eye, red eye, watery eyes with redness |
| Neuro | agitation, vision trouble, dizziness, confusion, delirium, agitated |
| Cardiac | palpitations, lightheadedness |
| Otitis | earache, ear ache, ear pain |
| Dermatitis | rash |

**Table S5. Synonym classification model performance**

| Symptom(s) | Total # sentences | % from template | Overall accuracy | True Pos Rate | False Neg Rate |
|---|---|---|---|---|---|
| COMBINED | 4001 | 10.2 | 0.874 | 0.927 | 0.167 |
| Fever / chills | 915 | 13.4 | 0.857 | 0.926 | 0.145 |
| Cough | 913 | 11.6 | 0.855 | 0.918 | 0.265 |
| Respiratory difficulty | 751 | 13.7 | 0.871 | 0.941 | 0.103 |
| Diarrhea | 440 | 17.3 | 0.899 | 0.913 | 0.15 |
| Respiratory failure | 202 | 0 | 0.985 | 0.995 | 0.5 |
| Headache | 199 | 0 | 0.852 | 0.942 | 0.115 |
| Fatigue | 198 | 0 | 0.914 | 0.971 | 0.25 |
| Altered or diminished sense of taste or smell | 166 | 0 | 0.728 | 0.644 | 0.361 |
| Myalgia/Arthralgia | 78 | 0 | 0.928 | 0.976 | 0.038 |
| Dysuria | 69 | 0 | 0.956 | 1 | 0 |
| Change in appetite/intake | 67 | 0 | 0.924 | 0.915 | 0.417 |
| Diaphoresis | 3 | 0 | 1 | 1 | 0 |

**Table S6. Performance of the BERT symptom classifier**

<u>18,490 Total Sentences</u>
- 11,505 tagged as YES for positive diagnosis
- 2265 tagged as NO for absence of disease/phenotype
- 2235 tagged as MAYBE for suspected disease/phenotype
- 2485 tagged as OTHER, e.g. for family history

**Overall Accuracy: 0.936** $= \dfrac{\text{\# of sentences the model labeled correctly in the test set}}{\text{Total \# of sentences in the test set}}$

Using a 90:10 split:

| Label | Precision | Recall | F1-Score |
|---|---|---|---|
| Other | 0.81 | 0.91 | 0.86 |
| Maybe | 0.87 | 0.83 | 0.85 |
| No | 0.97 | 0.95 | 0.96 |
| Yes | 0.97 | 0.96 | 0.96 |